\pdfoutput=1
\documentclass{article} 
\usepackage{iclr2027_conference,times}
\iclrfinalcopy

\usepackage{amsmath,amsfonts,bm}

\def\eqref#1{equation~\ref{#1}}

\def\1{\bm{1}}

\DeclareMathAlphabet{\mathsfit}{\encodingdefault}{\sfdefault}{m}{sl}
\SetMathAlphabet{\mathsfit}{bold}{\encodingdefault}{\sfdefault}{bx}{n}

\usepackage{hyperref}
\usepackage{url}
\usepackage{graphicx}
\usepackage{booktabs}
\usepackage{multirow}
\usepackage{amsmath}
\usepackage{amssymb}
\usepackage{xcolor}
\usepackage{colortbl}
\usepackage{enumitem}
\usepackage{pifont}
\usepackage{array}
\usepackage{tabularx}
\usepackage{float}
\usepackage{adjustbox}
\newcolumntype{L}[1]{>{\raggedright\arraybackslash}p{#1}}
\newcolumntype{Y}{>{\raggedright\arraybackslash}X}

\definecolor{passgreen}{HTML}{1B7837}
\definecolor{failred}{HTML}{B2182B}
\definecolor{mixorange}{HTML}{B35806}
\definecolor{rowgray}{HTML}{F2F2F2}
\definecolor{boxgray}{HTML}{F7F7F7}

\newcommand{\sig}{\textsuperscript{$\star$}}

\newcommand{\pp}{\,pp}
\newcommand{\SUP}{\textup{SUPPORTS}}
\newcommand{\REF}{\textup{REFUTES}}
\newcommand{\NEI}{\textup{NEI}}
\newcommand{\HIGH}{\textup{HIGH}}
\newcommand{\LOW}{\textup{LOW}}
\newcommand{\MED}{\textup{MEDIUM}}
\newcommand{\UNK}{\textup{UNKNOWN}}

\title{Multi-Channel Mitigation of Source-Trust\\ Shortcuts in Fact-Checking RL Agents}

\author{Jianchang Su \\ {\normalfont University of Connecticut}
\And Yiwei Yang \\ {\normalfont University of California, Santa Cruz}
\And Wei Zhang \\ {\normalfont University of Connecticut}}

\begin{document}

\maketitle
\lhead{Preprint}

\begin{abstract}
Retrieval-augmented fact-checkers often receive a reliability label, such as \HIGH{} or \LOW{} trust, for each evidence source. These labels should adjust the model's confidence and its decision to search for more evidence, while the verdict should follow the evidence content. We introduce \textsc{TrustSwap}, a counterfactual test that swaps, lowers, or removes source labels while keeping every evidence text fixed, and measures its three output channels (the verdict, the confidence, and the search decision) separately. Across untrained and RL-trained models at two scales, three datasets, and two prompts, confidence and search respond to the labels as intended in 49 of 50 comparisons, yet a label change alone alters 4--23\% of confident verdicts for Qwen3 models and up to 50\% for an existing RL-trained fact-checker. Standard GRPO fine-tuning amplifies this shortcut at 8B in all six settings. To reduce it, we propose trust-swap augmentation (TSA), which trains GRPO on each claim with both its original and its label-swapped evidence under the same gold verdict. At 4B, TSA lowers the verdict flip rate by 7--35\% (relative) in four of six settings, keeps accuracy and the intended confidence and search responses, outperforms reward-based alternatives in the main setting, and carries over to an unseen label-removal perturbation. An added consistency reward helps on the trained-on swap but not on unseen perturbations. At 8B, TSA's effect is not detectable, which makes scale the main open question.
\end{abstract}

\section{Introduction}
\label{sec:intro}

Automated fact-checking systems decide whether retrieved evidence supports or refutes a claim~\citep{guo2022survey,DBLP:conf/naacl/ThorneVCM18,DBLP:conf/nips/SchlichtkrullG023}. Recent systems train language models with reinforcement learning (RL) to search for evidence and answer~\citep{DBLP:journals/corr/abs-2503-09516}, including for claim verification~\citep{DBLP:journals/corr/abs-2510-01932}. The retrieved evidence comes from sources of very different reliability, so pipelines increasingly attach a reliability label to each source, for example from media-rating services~\citep{mbfc,ge2025confact}, and ask the model to take source credibility into account~\citep{pan2024credibility,deng2025cram}. This raises a basic question: \emph{what should a model do with these labels?}

\begin{figure}[t]
\centering
\includegraphics[width=\textwidth]{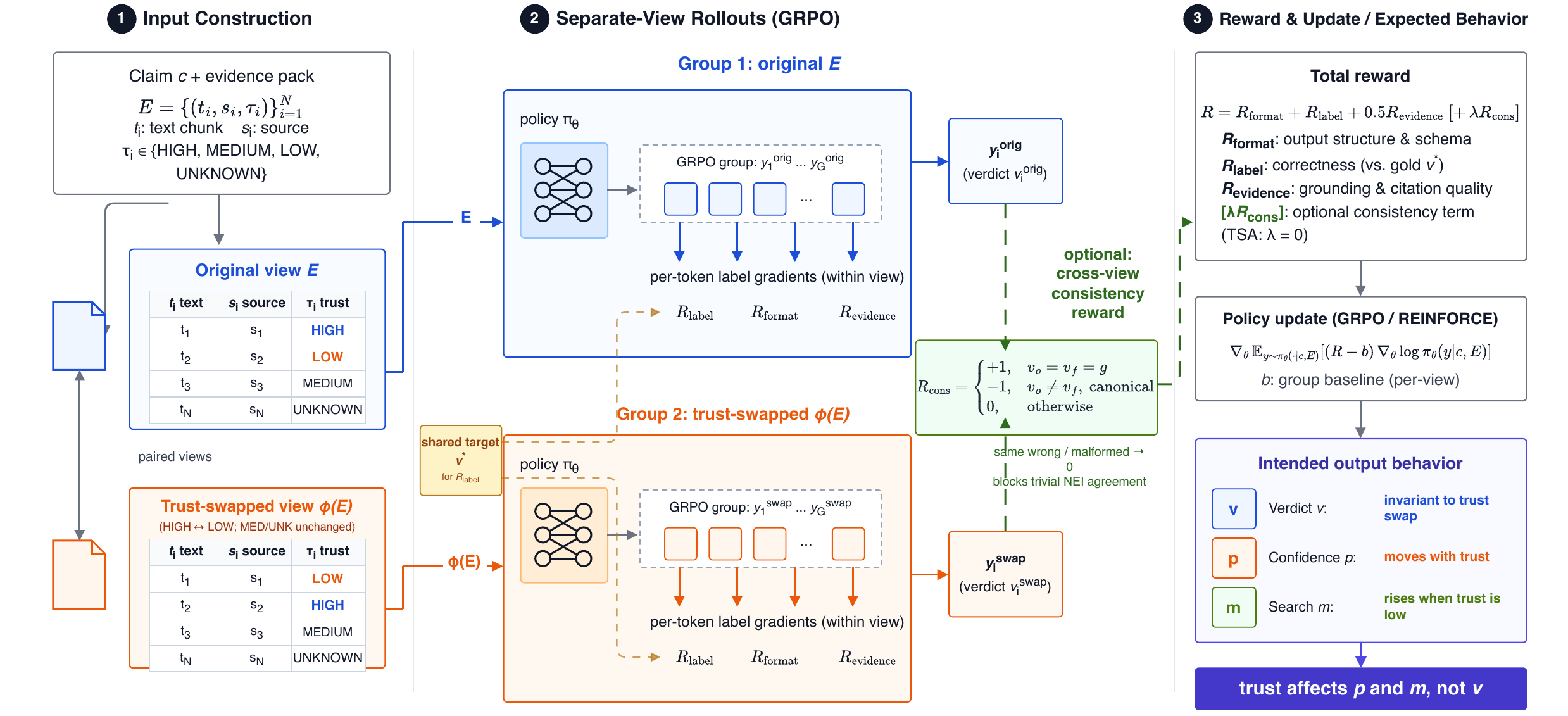}
\caption{\textsc{TrustSwap} and trust-swap augmentation (TSA). \textbf{(1)} Each claim is paired with its evidence pack $E$ and a trust-swapped pack $\phi(E)$ (\HIGH{}$\leftrightarrow$\LOW{}; texts and sources unchanged); \textsc{TrustSwap} compares the outputs on the two. \textbf{(2)} TSA trains GRPO on both packs as separate rollout groups with the same gold verdict $v^\star$; an optional consistency reward $R_{\text{cons}}$ links matched rollouts ($v_o$, $v_f$: verdicts on $E$ and $\phi(E)$; $g$: gold). \textbf{(3)} TSA uses $\lambda=0$. Trust should change the confidence $p$ and the search decision $m$, but not the verdict $v$.}
\label{fig:overview}
\end{figure}

We study a simple contract: \emph{the evidence content decides the verdict, and the labels adjust the model's confidence and its decision to look for more evidence.} Two considerations motivate it. First, reliability labels are metadata assigned by the pipeline, not properties of the evidence text. They can be wrong or outdated. Indirect prompt injection demonstrates manipulation of retrieved content~\citep{greshake2023injection}. Here, we assume that an attacker can change source labels while keeping the evidence text fixed. Second, the model already has two outputs for expressing doubt about sources, namely its confidence and its search decision, so keeping the verdict tied to the content loses no information. Other contracts are defensible; for example, a system could abstain whenever only low-reliability sources are available. We therefore also report verdict reversals separately from abstentions, so that our conclusions hold under this weaker contract as well.

To test the contract, we introduce \textsc{TrustSwap} (\S\ref{sec:trustswap}). For each claim, it exchanges the \HIGH{} and \LOW{} labels in the evidence pack, keeps every evidence text and source name fixed, and compares the model's outputs before and after the change (Figure~\ref{fig:overview}, left). Two further perturbations, which lower or remove the labels, test whether a model trained on the swap has learned something more general. Because the contract expects the verdict to stay fixed but confidence and search to move, \textsc{TrustSwap} measures these three output channels separately.

\textsc{TrustSwap} shows that the shortcut is specific to the verdict (\S\ref{sec:diagnosis}). Across five models, three datasets, and two prompts, \textbf{confidence and search already follow the contract} in 49 of 50 comparisons, so the models do read the labels as reliability signals. \textbf{The verdict does not follow the contract:} a label change alone alters 4--23\% of confident verdicts for Qwen3 models and 16--50\% for Veri-R1, an existing RL-trained fact-checker. \textbf{RL fine-tuning can amplify the shortcut:} GRPO training of Qwen3-8B raises the flip rate in all six settings, by 44\% in relative terms in the main setting, while accuracy in that setting changes by less than 2 percentage points (pp).

Because the failure is confined to the verdict, a fix should target the verdict and leave confidence and search free to respond. We propose \textbf{trust-swap augmentation (TSA)}, a counterfactual data augmentation~\citep{DBLP:conf/iclr/KaushikHL20} recipe for GRPO: each training claim appears twice, with its original and its label-swapped evidence, under the same gold verdict, and the two versions form separate rollout groups (Figure~\ref{fig:overview}; \S\ref{sec:method}). At 4B, TSA lowers the flip rate in four of six settings (by 7--35\% relative, all significant), keeps or slightly improves accuracy in all six, outperforms trust-aware reward shaping and a verdict-consistency reward in direct comparisons on the main setting, and carries over to an unseen perturbation that removes all labels. We also characterise where TSA stops helping: its gains come from fewer abstention changes rather than fewer reversals, and they do not appear at 8B.

Our contributions are:
\begin{itemize}[leftmargin=*,itemsep=1pt,topsep=2pt]
\item \textbf{\textsc{TrustSwap}}, a counterfactual test of source-label use: 1{,}584 claims from three datasets, three label perturbations, and separate measures for verdict (split into reversals and abstentions), confidence, and search (\S\ref{sec:trustswap}).
\item \textbf{A diagnosis across models and scales}: the source-label shortcut lives in the verdict, not in confidence or search, and standard GRPO fine-tuning amplifies it at 8B (\S\ref{sec:diagnosis}).
\item \textbf{Trust-swap augmentation}, a simple GRPO recipe that reduces the verdict shortcut at 4B without reducing accuracy, outperforms reward-based alternatives, and generalizes to an unseen perturbation, with its scope mapped at 8B (\S\ref{sec:method}--\S\ref{sec:results}).
\end{itemize}

\section{Related work}
\label{sec:related}

\paragraph{Fact-checking with retrieval and RL.}
Fact-checking benchmarks cover political claims, Wikipedia, science, and web evidence~\citep{DBLP:conf/acl/Wang17,DBLP:conf/naacl/ThorneVCM18,DBLP:conf/emnlp/JiangBZD0B20,DBLP:conf/nips/AlyGST00CM21,DBLP:conf/emnlp/WaddenLLWZCH20,DBLP:conf/nips/SchlichtkrullG023}, and retrieval-augmented models combine retrieved evidence with a language model~\citep{DBLP:conf/nips/LewisPPPKGKLYR020,DBLP:journals/corr/abs-2310-11511}. RL-trained agents learn to search and answer from reward signals~\citep{DBLP:journals/corr/abs-2510-01932,DBLP:journals/corr/abs-2503-09516,DBLP:journals/corr/abs-2402-03300}. We show that such training can increase reliance on source labels even when the reward never mentions them.

\paragraph{Source credibility and knowledge conflicts.}
Studies of knowledge conflicts ask how models weigh competing evidence and their own knowledge~\citep{xie2024chameleon,xu2024conflicts}, and credibility-aware methods give models reliability scores so that they rely less on unreliable context~\citep{pan2024credibility,deng2025cram,ge2025confact}. These works treat credibility as an input that should shape the answer; we ask \emph{which outputs} it should shape and find that the verdict is the one that should not change.

\paragraph{Calibration and abstention.}
Confidence calibration asks whether stated confidence matches accuracy~\citep{DBLP:conf/icml/GuoPSW17}, and selective prediction lets a model abstain when it is uncertain~\citep{DBLP:conf/icml/GeifmanE19}. In our setting, trust labels should move the confidence and the search decision but not the verdict, so an abstention caused only by a label change counts as a verdict change, which we report separately from reversals.

\paragraph{Shortcuts and counterfactual training.}
Shortcut learning, where models rely on features that correlate with the label but do not determine it, is well documented in NLP~\citep{DBLP:journals/natmi/GeirhosJMZBBW20,DBLP:conf/naacl/GururanganSLSBS18,DBLP:conf/acl/McCoyPL19}. Counterfactual and behavioural tests expose such shortcuts~\citep{gardner2020contrast,ribeiro2020checklist}, and counterfactual augmentation and consistency penalties are standard mitigations~\citep{DBLP:conf/iclr/KaushikHL20,garg2019counterfactual,veitch2021counterfactual}. In our setting the perturbed feature must be ignored by one output but used by the others, so we evaluate every method on all outputs at once.

\section{\textsc{TrustSwap}: a counterfactual test for source-label use}
\label{sec:trustswap}

\paragraph{Task and outputs.}
The input is a claim $c$ and an evidence pack $E=\{(t_i,s_i,\tau_i)\}_{i=1}^{N}$, where $t_i$ is the evidence text, $s_i$ the source domain, and $\tau_i\in\{\HIGH,\MED,\LOW,\UNK\}$ the source's trust label (we use ``trust label'', ``source label'', and ``reliability label'' interchangeably). The model answers in a fixed text format with a verdict $v\in\{\SUP,\REF,\NEI\}$, a confidence $p\in[0,1]$, a search flag $m\in\{0,1\}$ that states whether more evidence is needed, and a one-sentence rationale (prompts in Appendix~\ref{app:prompts}). The search flag is an output only; no second retrieval round is run.

\paragraph{Perturbations and pack types.}
The \emph{label swap} $\phi$ exchanges \HIGH{} and \LOW{} labels and leaves \MED{}, \UNK{}, every text $t_i$, and every source $s_i$ unchanged. It is the perturbation used for training in \S\ref{sec:method}. Two further perturbations are never used in training: a \emph{downshift} (\HIGH$\to$\MED, \MED$\to$\LOW) and \emph{label removal} (every label set to \UNK). The contract requires the verdict to stay fixed under all three. For the swap, the direction of the change depends on the pack. In a \emph{high-trust pack} (at least one \HIGH{} source and no \LOW{} source), the swap lowers trust; in a \emph{low-trust pack} (at least one \LOW{} source and no \HIGH{} source), it raises trust; in a \emph{mixed pack} (both), it reverses which sources are trusted, so no single direction is expected; \emph{unlabeled packs} (neither) are unchanged.

\paragraph{Metrics.}
We compare the original and perturbed predictions $(v,p,m)$ and $(v',p',m')$ on \emph{confident pairs}, where $\max(p,p')\ge 0.7$, because a verdict change matters most when the model commits to an answer. One measure follows each output. The \textbf{flip rate} is the percentage of confident pairs with $v\neq v'$; we split it into the \emph{reversal rate} (\SUP{}$\leftrightarrow$\REF{}) and the \emph{abstention-flip rate} (one of $v,v'$ is \NEI{}). The \textbf{confidence shift} is the mean of $p'-p$ and the \textbf{search shift} is the mean of $m'-m$, both in percentage points (pp) and reported per pack type. The contract predicts a flip rate of zero, a negative confidence shift and a positive search shift on high-trust packs, and the opposite signs on low-trust packs. To summarise each output in one number, the \emph{confidence response} is the pack-size-weighted mean of the confidence drop on high-trust packs and the confidence rise on low-trust packs, and the \emph{search response} is defined in the same way from the search shift; larger values mean a stronger response in the intended direction. Because decoding is greedy, every difference between the two predictions is caused by the label change.

\paragraph{Data.}
\textsc{TrustSwap} covers three datasets. AVeriTeC~\citep{DBLP:conf/nips/SchlichtkrullG023} provides 984 real-world claims with web evidence (700 high-trust, 158 low-trust, 85 mixed, and 41 unlabeled packs). CONFACT~\citep{ge2025confact} provides 400 claims whose sources often conflict (197 high-trust, 36 low-trust, 167 mixed). SciFact~\citep{DBLP:conf/emnlp/WaddenLLWZCH20} provides 200 claims over scientific abstracts, all in high-trust packs. Trust labels come from Media Bias/Fact Check ratings~\citep{mbfc}, a manual list of about 150 domains, and top-level-domain rules (Appendix~\ref{app:data}). We use two prompts. The \emph{neutral} prompt explains the labels without advice. The \emph{principled} prompt states the contract, and it also advises the model to request more search instead of committing to a verdict when only low-trust evidence is available. Our main setting is AVeriTeC with the neutral prompt, chosen because it has the most confident pairs (about 800).

\paragraph{Models.}
We test untrained Qwen3-4B-Instruct-2507 and Qwen3-8B (thinking disabled)~\citep{DBLP:journals/corr/abs-2505-09388}; Veri-R1, a Qwen2.5-3B model trained with online RL for claim verification~\citep{DBLP:journals/corr/abs-2510-01932}; and a \emph{GRPO baseline} for each Qwen3 size, trained as described in \S\ref{sec:method}. Evaluation uses greedy decoding.

\section{What models do with trust labels}
\label{sec:diagnosis}

\begin{figure}[t]
\centering
\small
\setlength{\tabcolsep}{4pt}
\renewcommand{\arraystretch}{1.12}
\begin{tabularx}{\textwidth}{@{}L{0.18\textwidth}Y L{0.13\textwidth}L{0.13\textwidth}@{}}
\toprule
\textbf{Claim} & \textbf{Evidence (source: label before $\rightarrow$ after the swap)} & \textbf{Original} & \textbf{Swapped} \\
\midrule
\multicolumn{4}{@{}p{0.98\textwidth}@{}}{\emph{(a) Typical case: labels lowered on a high-trust pack. Confidence and search respond as intended, but the verdict also changes.}}\\[1pt]
ICE removed thousands of gang members. &
[1] \texttt{ice.gov}: \HIGH{} $\rightarrow$ \LOW{}. ``5,396.''\newline
[2] \texttt{ice.gov}: \HIGH{} $\rightarrow$ \LOW{}. ``5,872.'' &
\SUP{}\newline conf.\ 0.95\newline search: no &
\textbf{\NEI{}}\newline conf.\ 0.3--0.5\newline search: yes \\
\midrule
\multicolumn{4}{@{}p{0.98\textwidth}@{}}{\emph{(b) Reversal: only one label is raised, and the verdict switches polarity with unchanged high confidence.}}\\[1pt]
Social distancing [\ldots] was not practiced at the funeral of civil rights icon John Lewis. &
[1] \texttt{perma.cc}: \UNK{}. ``\ldots the church will only accommodate 240 people in total, due to social distancing precautions.''\newline
[2] \texttt{facebook.com}: \LOW{} $\rightarrow$ \HIGH{}. ``Yes. The church's senior pastor \ldots\ referred to social distancing \ldots''\newline
[3] \texttt{gettyimages.com}: \UNK{}. ``Yes. Photo by \ldots'' &
\SUP{}\newline conf.\ 0.95\newline search: no &
\textbf{\REF{}}\newline conf.\ 0.95\newline search: no \\
\bottomrule
\end{tabularx}
\caption{The label swap on two AVeriTeC claims. Only the reliability labels change; every evidence text and source name is identical before and after. Outputs are from the 4B GRPO baseline, and both verdict changes occur in all five training seeds (the confidence range in (a) spans seeds). Under our contract, the confidence drop and the search request in (a) are correct, but neither verdict change is. The gold label is \SUP{} in (a) and \REF{} in (b).}
\label{fig:example}
\end{figure}

\begin{figure}[t]
\centering
\includegraphics[width=0.94\textwidth]{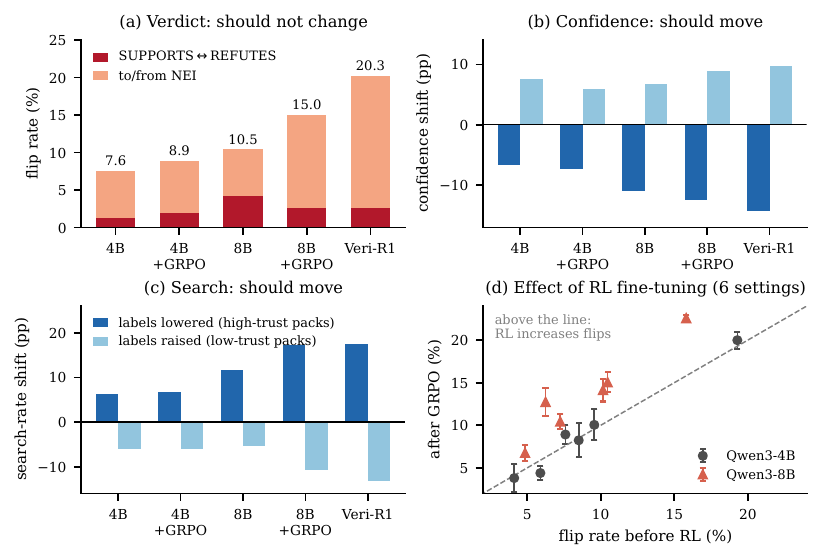}
\caption{\textsc{TrustSwap} label-swap results. \textbf{(a)--(c)} Main setting (AVeriTeC, neutral prompt): (a) flip rate, split into reversals and switches to or from \NEI{}; (b, c) mean change in confidence and search requests where labels are lowered (dark) or raised (light), both in the intended direction. \textbf{(d)} All six settings: flip rate before and after GRPO fine-tuning (mean $\pm$ s.d., five seeds); points above the diagonal mean that training increased flips.}
\label{fig:diagnosis}
\end{figure}

\paragraph{Confidence and search follow the contract.}
Figure~\ref{fig:example} shows two example cases, and Figure~\ref{fig:diagnosis} summarises all models. Figure~\ref{fig:diagnosis}b--c shows the two outputs that are supposed to respond to the labels. In the main setting, when the swap lowers trust, every model becomes less confident and asks for more search; when the swap raises trust, every model becomes more confident and asks for less search. The 4B GRPO baseline, for example, lowers its confidence by 7.3\pp{} and raises its search rate by 6.7\pp{} on high-trust packs, and moves both by about 6\pp{} in the opposite direction on low-trust packs. The pattern holds in 49 of 50 comparisons for each output (five models, both pack types, all datasets and prompts; Appendix~\ref{app:full-diagnosis}); the single exception is Veri-R1 on 10 low-trust CONFACT pairs. The models therefore use the labels for confidence and search as the contract intends.

\paragraph{The verdict does not follow the contract.}
The same swap also changes the verdict (Figure~\ref{fig:diagnosis}a). In the main setting, the flip rate is 7.6\% for untrained Qwen3-4B, 8.9\% for its GRPO baseline, 10.5\% for untrained Qwen3-8B, 15.0\% for its GRPO baseline, and 20.3\% for Veri-R1. Across all six settings, it ranges from 3.8\% to 22.6\% for the Qwen3 models and from 15.8\% to 50.0\% for Veri-R1. Most flips are abstention changes (78\% for the 4B baseline): 73\% of these are switches \emph{to} \NEI{}, and 95\% of those occur on high-trust packs, where the swap lowers trust (Figure~\ref{fig:example}a). A weaker contract could accept such abstentions, but reversals cannot be justified in this way, and they are not rare: they make up 13--40\% of flips in the main setting. For the 4B baseline, 74\% of reversals occur on high-trust packs, where the swap lowers every \HIGH{} source together and leaves the evidence text unchanged, so no reliability-weighted reading of the evidence explains a switch between \SUP{} and \REF{}. Figure~\ref{fig:example}b shows a reversal caused by raising a single label.

\paragraph{Stating the rule in the prompt does not help.}
The principled prompt, which states the contract, raises the flip rate in 14 of 15 model--dataset pairs, for example from 8.9\% to 20.0\% for the 4B baseline on AVeriTeC. Almost all of this is abstention (18.6 of 20.0 points), in line with the prompt's advice not to commit when only low-trust evidence is available. Prompting alone is therefore not enough, which motivates a training-time fix.

\paragraph{RL fine-tuning can amplify the shortcut.}
Figure~\ref{fig:diagnosis}d compares each GRPO baseline with its untrained model. At 8B, training raises the flip rate in every setting, by 1.9--6.8\pp{}, and all 30 seed--setting runs lie above the untrained value; in the main setting the rate rises from 10.5\% to 15.0\%, while accuracy stays close (67.0\% before and $65.1\pm0.6$\% after). The increase comes from abstention flips, which double from 6.3\% to 12.5\% of confident pairs, while reversals fall from 4.2\% to 2.6\%. At 4B the change is small and mixed ($-1.5$ to $+1.3$\pp{} across settings). The reward contains no trust term, so this behaviour is learned indirectly; an outcome reward alone therefore does not protect against the shortcut.

\section{Trust-swap augmentation}
\label{sec:method}

\paragraph{GRPO baseline.}
We fine-tune each Qwen3 model with Group Relative Policy Optimization (GRPO)~\citep{DBLP:journals/corr/abs-2402-03300} and LoRA of rank 32~\citep{DBLP:conf/iclr/HuSWALWWC22} for 25 steps, with 16 claims per step and $G{=}8$ rollouts per claim. The reward is $R=R_{\text{format}}+R_{\text{label}}+0.5\,R_{\text{evidence}}$: a format term, a label-correctness term, and an evidence-citation term, none of which refers to trust labels. Training uses 311 claims from CONFACT and SciFact, excluding evaluation claims by identifier (\REF{} 67.8\%, \SUP{} 17.7\%, \NEI{} 14.5\%, resampled to 40/30/30). One CONFACT claim text appears in both sets. Source labels and class labels are correlated in this pool: every training claim with a \LOW{}-trust source is labelled \REF{}. The learning rate is $4\times10^{-5}$ at 4B and $10^{-5}$ at 8B (\S\ref{sec:protocol}); Appendix~\ref{app:details} lists all settings. This short run is a controlled instance of RL fine-tuning for fact-checking, not an attempt to maximise accuracy; AVeriTeC is out of domain for the training claims.

\paragraph{Trust-swap augmentation (TSA).}
The diagnosis shows that the model should become \emph{invariant} to the labels in its verdict but stay \emph{sensitive} to them in its confidence and search. TSA gives the verdict a direct training signal for this invariance (Figure~\ref{fig:overview}). For every training claim whose pack contains a \HIGH{} or \LOW{} label, we add the swapped pack $\phi(E)$ with the same gold verdict. The original and swapped versions form two separate GRPO groups of $G$ rollouts each, so each version's advantages are normalised within its own group and the label-correctness reward pushes both versions toward the same gold verdict. The reward is unchanged, and no term constrains confidence or search, which remain free to respond to the labels. TSA doubles the number of rollouts per claim but changes nothing else in the baseline recipe.

\paragraph{Optional consistency reward.}
We also test an explicit link between the two versions. For matched rollouts on $E$ and $\phi(E)$ with verdicts $v$ and $v'$ and gold verdict $g$, the consistency reward is $R_{\text{cons}}=+1$ if $v=v'=g$, $-1$ if $v\neq v'$ and both are well-formed, and $0$ otherwise; the zero case stops the model from earning the reward by answering \NEI{} on both. \emph{TSA + consistency} adds $\lambda R_{\text{cons}}$ to the reward of both groups.

\paragraph{Baselines.}
Besides the GRPO baseline, we compare with two families that also target trust behaviour. \emph{Trust-aware reward shaping} adds four reward terms that ask confidence and search to track source trust, confidence to match correctness, and the verdict to agree with an NLI model when the NLI signal is clear, with an extra penalty for disagreeing with the NLI model on packs that contain \LOW{} sources; a second variant adds a denser reward for abstaining on weakly supported claims. The \emph{consistency reward} alone also samples rollouts on both $E$ and $\phi(E)$, but places them in one joint group of $2G$ rollouts and adds $\lambda R_{\text{cons}}$, for $\lambda\in\{0.1,0.2,0.4\}$. It therefore differs from TSA + consistency only in using a joint rather than a separate group per version, and it samples as many rollouts per claim as TSA, so comparisons with TSA are matched in compute.

\subsection{Evaluation protocol}
\label{sec:protocol}

\paragraph{Seeds and statistics.}
We train five seeds for the GRPO baseline and trust-aware rewards, and three for the other arms at 4B; at 8B, all arms have five matched seeds. Because single-seed RL results are often unreliable~\citep{DBLP:conf/aaai/0002IBPPM18,bouthillier2021variance}, we report differences in percentage points with 95\% bootstrap intervals over seeds (10{,}000 resamples)~\citep{DBLP:conf/nips/AgarwalSCCB21} and mark with $\star$ intervals that exclude zero. Comparisons against the 4B baseline resample seeds independently within each arm; comparisons between two arms that share seed indices resample the per-seed differences. With three seeds an interval excludes zero essentially only when all per-seed differences share a sign (one-sided sign-test $p=0.125$), so we confirm 4B results with the unseen perturbations and the 8B replication. The main setting was fixed before the experiments; we do not correct for multiple comparisons, and results in the other five settings serve as supporting evidence.

\paragraph{8B training.}
The 4B learning rate made 8B training unstable (mean accuracy fell to 51--59\% across arms; Appendix~\ref{app:8b}). We therefore re-selected the learning rate on the baseline arm only, requiring accuracy within 2\pp{} of the untrained model and an \NEI{} rate of at most 30\%, without looking at any \textsc{TrustSwap} metric. Of three candidates ($2\times10^{-5}$ and $10^{-5}$ for 25 steps, $4\times10^{-5}$ for 10 steps), only $10^{-5}$ passed. We moved to 8B rather than adding 4B seeds because the hosting service withdrew the 4B base model from training during the study.

\section{Results}
\label{sec:results}

\begin{table}[t]
\centering
\small
\caption{Change in flip rate (pp) relative to the 4B GRPO baseline in all six settings (N/P: neutral/principled prompt); first row: baseline flip rate (\%, mean $\pm$ s.d., five seeds). Negative is better. $\star$: 95\% interval excludes zero. Last row: TSA + consistency ($\lambda{=}0.4$) minus TSA on matched seeds.}
\label{tab:main}
\begin{adjustbox}{max width=\textwidth}\begin{tabular}{@{}lcccccc@{}}
\toprule
Method & AVeriTeC N & AVeriTeC P & SciFact N & SciFact P & CONFACT N & CONFACT P \\
\midrule
GRPO baseline (\%) & $8.92\pm1.11$ & $19.99\pm0.99$ & $3.80\pm1.65$ & $10.05\pm1.82$ & $4.40\pm0.79$ & $8.24\pm1.98$ \\
\midrule
Trust-aware reward & $+$0.76 & $+$0.23 & $-$0.32 & $+$1.28 & $+$0.05 & $+$1.05 \\
Trust-aware + abstention & $-$0.12 & $+$0.28 & $+$0.77 & $-$0.36 & $+$0.44 & $-$0.52 \\
\midrule
Consistency, $\lambda{=}0.1$ & $+$0.46 & $-$0.12 & $+$0.84 & $-$2.08 & $-$0.86\sig & $-$0.55 \\
Consistency, $\lambda{=}0.2$ & $-$0.19 & $-$1.14 & $+$0.28 & $-$2.22 & $-$0.67 & $-$0.37 \\
Consistency, $\lambda{=}0.4$ & $+$0.79 & $-$0.61 & $-$0.25 & $-$1.28 & $-$1.09\sig & $-$1.69 \\
\midrule
\rowcolor{rowgray} \textbf{TSA (ours)} & $-$1.12\sig & $-$1.49\sig & $-$0.56 & $+$0.44 & $-$1.52\sig & $-$1.66\sig \\
\rowcolor{rowgray} TSA + consistency, $\lambda{=}0.2$ & $-$1.15\sig & $-$0.43 & $+$0.50 & $-$0.48 & $-$0.32 & $-$0.75 \\
\rowcolor{rowgray} TSA + consistency, $\lambda{=}0.4$ & $-$1.60\sig & $-$0.99\sig & $-$0.51 & $-$1.00 & $-$1.03 & $-$0.93 \\
\midrule
TSA + consistency ($\lambda{=}0.4$) $-$ TSA & $-$0.48 & $+$0.50 & $+$0.05 & $-$1.44\sig & $+$0.50 & $+$0.74 \\
\bottomrule
\end{tabular}
\end{adjustbox}
\end{table}

\paragraph{TSA reduces verdict flips.}
Table~\ref{tab:main} reports every method in all six settings. TSA lowers the flip rate significantly in four settings: by 1.12\pp{} (13\% relative) in the main setting, 1.49\pp{} (7\%) on AVeriTeC with the principled prompt, and 1.52\pp{} (35\%) and 1.66\pp{} (20\%) on CONFACT. Accuracy is maintained: TSA's accuracy is equal to or slightly higher than the baseline's in all six settings ($+0.2$ to $+1.2$\pp; Table~\ref{tab:accuracy} in Appendix~\ref{app:breakdown}). TSA + consistency ($\lambda{=}0.4$) gives the largest reduction in the main setting ($-1.60$\pp, 18\% relative) and a lower mean flip rate than the baseline in all six settings, two of them significantly. The result does not hinge on the 0.7 confidence threshold: TSA's estimate is negative at every threshold from 0 to 0.9 and significant from 0.5 to 0.7 (Appendix~\ref{app:threshold}).

\paragraph{TSA outperforms reward-based alternatives.}
Neither trust-aware reward changes the flip rate in the main setting ($+0.76$ and $-0.12$\pp, both intervals include zero), and the consistency reward alone does not reduce it at any weight. Direct comparisons confirm the gap: in the main setting, TSA's flip rate is lower than that of the trust-aware reward by $1.86$\pp\sig{}, of the trust-aware reward with abstention by $0.99$\pp\sig{}, and of the consistency reward by $1.58$\pp\sig{} ($\lambda{=}0.1$) and $1.91$\pp\sig{} ($\lambda{=}0.4$; at $\lambda{=}0.2$ the difference, $-0.92$\pp, is not significant). Across the six settings, TSA is significantly better than the trust-aware reward in five. TSA is not better everywhere: on SciFact with the principled prompt, the consistency reward alone gives a lower flip rate. Overall, supervising the verdict directly on both versions of a claim is more effective than rewarding trust-appropriate behaviour.

\begin{table}[t]
\centering
\small
\caption{All outputs at 4B in the main setting (AVeriTeC, neutral prompt). $\Delta$ columns are differences from the GRPO baseline. The verdict columns should decrease: flip rate and its two parts, reversals and abstention flips (pp). The trust-response columns should \emph{not} decrease: the confidence response and the search response measure how strongly confidence and search move in the intended direction under the swap (pp; baseline values in the first row). $\star$: 95\% interval excludes zero.}
\label{tab:breakdown}
\begin{adjustbox}{max width=\textwidth}\begin{tabular}{@{}lcccccccc@{}}
\toprule
 & \multicolumn{4}{c}{Verdict (should not change)} & \multicolumn{2}{c}{Trust response (should stay)} & \multicolumn{2}{c}{Quality} \\
\cmidrule(lr){2-5}\cmidrule(lr){6-7}\cmidrule(lr){8-9}
Method & Flip (\%) & $\Delta$ flip [95\% CI] & $\Delta$ rev. & $\Delta$ abst. & $\Delta$ conf. & $\Delta$ search & Acc.\ (\%) & \NEI{} (\%) \\
\midrule
GRPO baseline & $8.92\pm1.11$ & -- & -- & -- & 7.03 & 6.62 & $61.7\pm2.3$ & $31.0\pm5.0$ \\
\midrule
Trust-aware reward & $9.67\pm1.31$ & $+$0.76 [$-$0.61, $+$2.06] & $+$0.44 & $+$0.32 & $+$0.26 & $+$0.44 & $62.5\pm2.5$ & $30.5\pm5.8$ \\
Trust-aware + abstention & $8.80\pm0.32$ & $-$0.12 [$-$1.06, $+$0.74] & $+$0.17 & $-$0.29 & $+$0.19 & $-$0.21 & $63.4\pm2.2$ & $29.1\pm4.5$ \\
\midrule
Consistency, $\lambda{=}0.1$ & $9.39\pm1.60$ & $+$0.46 [$-$1.10, $+$2.25] & $+$0.70\sig & $-$0.24 & $+$1.07\sig & $+$0.36 & $61.8\pm0.2$ & $30.9\pm0.4$ \\
Consistency, $\lambda{=}0.2$ & $8.73\pm1.05$ & $-$0.19 [$-$1.53, $+$1.03] & $+$0.14 & $-$0.34 & $+$0.23 & $-$0.19 & $61.4\pm1.8$ & $31.0\pm4.0$ \\
Consistency, $\lambda{=}0.4$ & $9.71\pm1.17$ & $+$0.79 [$-$0.63, $+$2.10] & $+$0.76 & $+$0.03 & $+$0.94\sig & $-$0.20 & $61.8\pm2.0$ & $30.3\pm4.5$ \\
\midrule
\rowcolor{rowgray} \textbf{TSA (ours)} & $7.80\pm0.15$ & $-$1.12\sig [$-$2.04, $-$0.32] & $-$0.01 & $-$1.11\sig & $-$0.67 & $-$0.36 & $62.3\pm0.7$ & $30.0\pm1.7$ \\
\rowcolor{rowgray} TSA + consistency, $\lambda{=}0.2$ & $7.77\pm0.82$ & $-$1.15\sig [$-$2.32, $-$0.05] & $-$0.25 & $-$0.90 & $+$0.25 & $-$0.27 & $61.8\pm1.2$ & $33.4\pm3.1$ \\
\rowcolor{rowgray} TSA + consistency, $\lambda{=}0.4$ & $7.32\pm1.13$ & $-$1.60\sig [$-$2.90, $-$0.18] & $-$0.22 & $-$1.38\sig & $-$0.76 & $-$1.00\sig & $61.2\pm1.0$ & $32.4\pm3.2$ \\
\bottomrule
\end{tabular}
\end{adjustbox}
\end{table}

\paragraph{TSA keeps the trust response that the contract asks for.}
A method could reduce flips by making the model ignore the labels altogether, which would also remove the useful confidence and search responses. Table~\ref{tab:breakdown} shows that TSA does not do this. Its confidence response ($-0.67$) and search response ($-0.36$) do not change significantly, and accuracy ($62.3$\% vs.\ $61.7$\%) and the \NEI{} rate ($30.0$\% vs.\ $31.0$\%) stay close to the baseline. The reduction is concentrated on high-trust packs ($-1.29$\pp\sig), where a verdict change is hardest to justify, with a smaller reduction on low-trust packs ($-1.03$\pp) and none on mixed packs ($-0.12$\pp). In contrast, TSA + consistency ($\lambda{=}0.4$) weakens the search response ($-1.00$\sig), another reason to prefer plain TSA.

\begin{figure}[t]
\centering
\includegraphics[width=0.8\textwidth]{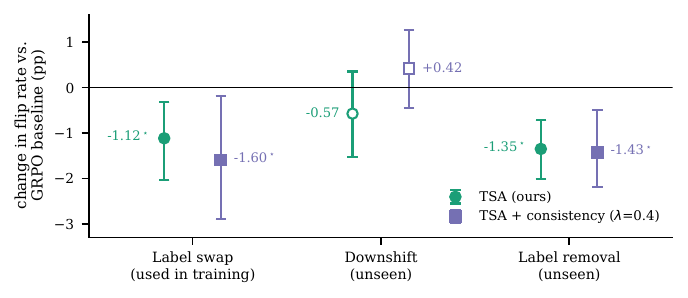}
\caption{Change in flip rate relative to the 4B GRPO baseline (main setting) under the label swap used in training and under two label perturbations never seen in training. Filled markers: 95\% interval excludes zero. TSA's gain carries over to label removal and keeps its direction under the downshift.}
\label{fig:checks}
\end{figure}

\paragraph{TSA generalizes to unseen perturbations.}
A method trained on the label swap might only learn to ignore that particular change. Figure~\ref{fig:checks} evaluates TSA on the two perturbations that it never saw. Under label removal, TSA lowers the flip rate by $1.35$\pp\sig{} ($[-2.01,-0.72]$), slightly more than on the swap it was trained on. Under the downshift, the reduction keeps its direction ($-0.57$\pp, $[-1.52,+0.36]$). TSA therefore meets a strict criterion for generalization: a reduction under both unseen perturbations, significant for at least one (details in Appendix~\ref{app:unseen}). TSA + consistency also carries over to label removal ($-1.43$\pp\sig) but not to the downshift, where it is $0.99$\pp\sig{} worse than TSA ($[+0.74,+1.37]$, three of three seeds). The explicit consistency term thus buys a larger gain on the trained-on perturbation at the cost of robustness to new ones, which is why we recommend plain TSA.

\paragraph{What each component changes.}
The flip-type columns of Table~\ref{tab:breakdown} show how the methods differ. TSA removes abstention flips ($-1.11$\pp\sig, 16\% relative) while its confidence response stays unchanged: when \HIGH{} sources become \LOW{}, the TSA model more often keeps its verdict and still lowers its confidence, moving the expression of doubt from the verdict to the confidence as the contract asks. Reversals are unchanged ($-0.01$\pp); no method in our study significantly reduces reversals under the swap. Nor does TSA fix the persistent failures: the 29 claims (3\%) whose verdict flips in at least four of five baseline runs still flip under TSA, 23 of them in all three runs, so its gain comes from claims whose verdict flips in only some runs (Appendix~\ref{app:claims}). The consistency reward acts mainly on confidence: it strengthens the confidence response by 15\% ($+1.07$\sig{} at $\lambda{=}0.1$) and 13\% ($+0.94$\sig{} at $\lambda{=}0.4$), while slightly increasing reversals at $\lambda{=}0.1$ ($+0.70$\pp\sig). Group structure matters: with the same consistency reward and compute, separate groups instead of one joint group lower the flip rate by $2.38$\pp\sig{} at $\lambda{=}0.4$ ($-0.96$\pp, not significant, at $\lambda{=}0.2$).

\begin{table}[t]
\centering
\small
\caption{Qwen3-8B in the main setting (five matched seeds; learning rate re-selected on the baseline, \S\ref{sec:protocol}). $\Delta$ flip compares each method with the 8B GRPO baseline on matched seeds. All six settings are in Appendix~\ref{app:8b}.}
\label{tab:8b-main}
\begin{adjustbox}{max width=\textwidth}\begin{tabular}{@{}lcccccc@{}}
\toprule
Qwen3-8B & Flip rate (\%) & $\Delta$ flip [95\% CI] & Rev.\ / abst.\ (\%) & Conf.\ resp. & Acc.\ (\%) & \NEI{} (\%) \\
\midrule
Untrained & 10.5 & -- & 4.2 / 6.3 & 10.2 & 67.0 & 16.0 \\
GRPO baseline & $15.0\pm1.2$ & -- & 2.6 / 12.5 & 11.8 & $65.1\pm0.6$ & $26.0\pm1.0$ \\
\rowcolor{rowgray} TSA (ours) & $14.7\pm1.6$ & $-$0.36 [$-$2.21, $+$1.33] & 3.0 / 11.7 & 12.3 & $64.2\pm0.8$ & $24.7\pm1.7$ \\
\rowcolor{rowgray} TSA + consistency, $\lambda{=}0.4$ & $14.3\pm1.3$ & $-$0.74 [$-$2.49, $+$0.58] & 3.2 / 11.1 & 11.8 & $64.4\pm0.3$ & $24.1\pm1.3$ \\
\bottomrule
\end{tabular}
\end{adjustbox}
\end{table}

\paragraph{Scaling to 8B.}
Table~\ref{tab:8b-main} repeats the main comparison on Qwen3-8B. The diagnosis carries over and becomes stronger: the 8B models have a larger confidence response than the 4B models (10.2--12.3 vs.\ about 7), and GRPO training raises the flip rate from 10.5\% to 15.0\%, entirely through abstention flips. TSA keeps the confidence response and accuracy at 8B, but its reduction in flips is not detectable ($-0.36$\pp, $[-2.21,+1.33]$; TSA + consistency $-0.74$\pp, $[-2.49,+0.58]$) in any setting (Appendix~\ref{app:8b}), and both methods slightly increase reversals ($+0.39$\pp\sig{} and $+0.66$\pp\sig). At this scale, the amplification caused by GRPO itself ($+4.6$\pp) is larger than the effect of either mitigation at our training budget. One untested explanation is that the 8B runs use a four times smaller learning rate for the same 25 steps, which weakens the extra signal from the swapped groups.

\section{Discussion}
\label{sec:discussion}

\paragraph{Recommendations for evaluation.}
Measuring each output separately showed that the shortcut is confined to the verdict and that the consistency reward changes confidence rather than the verdict. Testing on unseen perturbations separated the method that generalizes (TSA) from the one that mainly fits the training perturbation (TSA + consistency). This result motivates evaluation on perturbations not used during training. Contrast sets provide a related approach to testing model behaviour beyond the original test set~\citep{gardner2020contrast}.

\paragraph{Choosing a contract.}
Whether abstaining under low trust is acceptable depends on the deployment. A system that forwards abstentions to human fact-checkers may prefer the weaker contract, in which only reversals are errors; a system whose verdicts are published directly, or whose labels can be manipulated, needs the stricter one. Because \textsc{TrustSwap} reports reversals and abstentions separately, it supports both choices. Under the stricter contract, TSA removes a sixth of label-driven abstentions at 4B; under either contract, reversals remain the open problem.

\paragraph{Toward larger models.}
At 8B, GRPO itself amplifies the shortcut and TSA does not offset it within our 25-step, 311-claim budget. Natural next steps are longer training on more claims, training data in which labels and verdicts are not correlated, and designs that keep labels out of the verdict computation while passing them to the confidence and search outputs.

\paragraph{Limitations.}
We study one model family at two sizes, one RL algorithm with short LoRA training, and a single-pass pipeline in which the search flag is not executed. The unseen perturbations were run at 4B in the main setting only, and new 4B arms have three seeds. The 8B evaluation used the hosted sampling service and the 4B evaluation used local inference, so we compare methods only within a model size. Trust labels come from heuristic domain ratings, and the training pool correlates \LOW{} labels with \REF{}. Veri-R1 is evaluated with our prompt format rather than its own pipeline, so its numbers show that the shortcut exists, not how it performs in deployment. Our metrics are behavioural and have not been validated against human judgements. Finally, the contract is a design choice, which is why we report reversals separately.

\section{Conclusion}
\label{sec:conclusion}

\textsc{TrustSwap} shows that fact-checking models use source-reliability labels correctly for confidence and search, but also let them decide the verdict, and that GRPO fine-tuning can amplify this. Trust-swap augmentation reduces the shortcut at 4B without hurting accuracy or the useful trust responses, outperforms reward-based alternatives, and generalizes to an unseen perturbation. Extending these gains to larger models and to verdict reversals is the main open problem that \textsc{TrustSwap} makes measurable.

\bibliography{references}
\bibliographystyle{iclr2027_conference}

\newpage
\appendix

\section*{Ethics statement}
This work studies how fact-checking models respond to source-reliability labels, with the goal of making such systems harder to manipulate. The label perturbations could in principle guide an attacker who controls labels; we report aggregate behaviour only and release no attack tooling beyond the perturbation scripts needed for replication. All datasets are public research datasets. Source-reliability ratings encode judgements that can themselves be contested, which is one reason we argue that they should not decide verdicts.

\section*{Reproducibility statement}
Our code includes the diagnostic construction scripts, the label perturbations, inference and metric code, reward and training code (including TSA and the consistency reward), prompts, and unit tests. Appendix~\ref{app:details} lists all hyperparameters. All numbers and figures are computed from per-example prediction files by stdlib-only scripts. We will release this code together with all trained LoRA adapters, prediction files, and analysis scripts. This matters because the 4B base model is no longer available for training on the hosted service we used, so released adapters and predictions are the stable way to reproduce the 4B results.

\section*{AI use statement}
We used AI assistants for writing support only, namely grammar correction, wording, and proofreading. The authors reviewed all text and take full responsibility for the content of the paper.

\section{Experimental details}
\label{app:details}

\paragraph{Training.}
All arms use LoRA of rank 32, GRPO with 16 claims per step and $G{=}8$ rollouts per group, 25 steps, a KL penalty of 0.1 towards the base model, sampling temperature 1.3, a maximum response length of 256 tokens, and class resampling of 40/30/30 over \REF/\SUP/\NEI. The learning rate is $4\times10^{-5}$ at 4B and $1\times10^{-5}$ at 8B. The base reward is $R=R_{\text{format}}+R_{\text{label}}+0.5\,R_{\text{evidence}}$. The trust-aware reward adds a trust-behaviour term (weight 0.3; confidence, search, and conflict handling should track source trust and evidence sufficiency), a calibration term (0.2; $1-|p-\mathbb{1}[\text{correct}]|$), an entailment term (0.4; the verdict should match an NLI model's prediction, strictly when the NLI signal is clear and loosely otherwise), and a penalty (0.3) when the verdict disagrees with the NLI prediction on packs that contain \LOW{}-trust sources. None of these terms uses the swapped pack. The abstention variant adds a denser reward for abstaining on claims whose evidence is weak according to the NLI model. In TSA, each swapped version is an additional GRPO group with its own advantage normalisation. In TSA + consistency, the two groups of a claim exchange $R_{\text{cons}}$ after both have been sampled. The consistency reward alone uses one joint group of $2G$ rollouts per claim. Training ran on the Tinker hosted service; a 25-step 4B run cost about US\$15.

\paragraph{Evaluation.}
4B models were evaluated with local Hugging Face inference on one GPU; 8B models with the hosted sampling API. Both use greedy decoding with at most 256 new tokens. For 8B, we checked that the hosted prompt renderer produces byte-identical prompts to the local chat template with thinking disabled. None of the 478{,}325 prediction records in our evaluation files lacks a valid verdict.

\paragraph{Training data.}
The 311 training claims are 211 CONFACT claims (all \REF) and 100 SciFact claims, selected to exclude every evaluation claim by identifier (one CONFACT claim text appears in both sets). Every training claim with a \LOW{}-trust source is labelled \REF{}, and every \SUP{} and \NEI{} claim has only \HIGH{}-trust sources.

\section{Data construction and prompts}
\label{app:data}
\label{app:prompts}

Trust labels are assigned per source domain: a manual list of about 150 domains takes precedence (peer-reviewed journals, government, and wire services as \HIGH; tabloids and social media as \LOW; mainstream news as \MED), followed by Media Bias/Fact Check credibility ratings (about 6{,}100 domains, distributed with the CONFACT repository) and top-level-domain rules for unlisted \texttt{.gov} and \texttt{.edu} domains. Other domains are \UNK. The AVeriTeC evidence is built from the question--answer evidence of each claim, keeping claims with at least two evidence items and at least one labelled source. The perturbed packs are identical to the original except for the labels.

\noindent\textbf{Neutral prompt.}
{\small\ttfamily You are a fact-checking assistant. Given a claim and a list of evidence pieces (with their source domains and a `trust' label per source), determine whether the claim is SUPPORTED by the evidence, REFUTED by it, or has Not Enough Information (NEI). [claim] [evidence] Respond using EXACTLY this format: verdict / confidence / needs\_more\_search / evidence\_ids\_used / reasoning.}

\noindent\textbf{Principled prompt} adds: {\small\ttfamily The verdict is determined by the EVIDENCE CONTENT. The `trust' label is NOT a truth signal. A HIGH-trust source can be wrong; a LOW-trust source can be right. Do not let the trust label flip the verdict. Use the `trust' label to CALIBRATE your confidence \ldots\ Use the `trust' label to decide needs\_more\_search: if the only evidence supporting/refuting the claim comes from low-trust sources, set needs\_more\_search: yes rather than committing to a definitive verdict. If evidence is weak or absent regardless of trust, prefer NEI over SUPPORTS/REFUTES.}

\section{Full diagnosis}
\label{app:full-diagnosis}

\begin{table}[H]
\centering
\small
\setlength{\tabcolsep}{3.3pt}
\caption{Label-swap results for all models and settings. Flip: flip rate (\%); Rev.: reversal rate (\% of confident pairs); $\Delta p$ and $\Delta m$: confidence and search shift (pp) on high-trust (H) and low-trust (L) packs. GRPO rows are means over five seeds. SciFact has only high-trust packs.}
\label{tab:full-diagnosis}
\begin{adjustbox}{max width=\textwidth}\begin{tabular}{@{}llrrrrrr@{}}
\toprule
Setting & Model & Flip & Rev. & $\Delta p$ H & $\Delta p$ L & $\Delta m$ H & $\Delta m$ L \\
\midrule
AVeriTeC N & Qwen3-4B & 7.6 & 1.2 & $-$6.6 & $+$7.6 & $+$6.2 & $-$6.1 \\
 & Qwen3-4B + GRPO & 8.9 & 1.9 & $-$7.3 & $+$6.0 & $+$6.7 & $-$6.1 \\
 & Qwen3-8B & 10.5 & 4.2 & $-$11.0 & $+$6.7 & $+$11.7 & $-$5.3 \\
 & Qwen3-8B + GRPO & 15.0 & 2.6 & $-$12.4 & $+$8.9 & $+$17.3 & $-$10.6 \\
 & Veri-R1 & 20.3 & 2.6 & $-$14.3 & $+$9.8 & $+$17.6 & $-$13.1 \\
\addlinespace[2pt]
AVeriTeC P & Qwen3-4B & 19.3 & 1.6 & $-$24.5 & $+$22.8 & $+$19.6 & $-$20.6 \\
 & Qwen3-4B + GRPO & 20.0 & 1.4 & $-$25.1 & $+$22.8 & $+$22.2 & $-$22.2 \\
 & Qwen3-8B & 15.8 & 2.3 & $-$27.1 & $+$25.2 & $+$73.6 & $-$73.6 \\
 & Qwen3-8B + GRPO & 22.6 & 2.3 & $-$27.6 & $+$26.3 & $+$73.7 & $-$75.8 \\
 & Veri-R1 & 25.0 & 6.0 & $-$13.3 & $+$14.2 & $+$18.5 & $-$20.6 \\
\addlinespace[2pt]
SciFact N & Qwen3-4B & 4.1 & 0.0 & $-$5.5 & -- & $+$5.9 & -- \\
 & Qwen3-4B + GRPO & 3.8 & 0.6 & $-$4.2 & -- & $+$3.4 & -- \\
 & Qwen3-8B & 6.2 & 3.1 & $-$5.8 & -- & $+$3.1 & -- \\
 & Qwen3-8B + GRPO & 12.7 & 1.9 & $-$7.9 & -- & $+$10.8 & -- \\
 & Veri-R1 & 15.8 & 4.8 & $-$11.1 & -- & $+$11.0 & -- \\
\addlinespace[2pt]
SciFact P & Qwen3-4B & 9.6 & 0.7 & $-$19.7 & -- & $+$8.8 & -- \\
 & Qwen3-4B + GRPO & 10.1 & 0.8 & $-$21.3 & -- & $+$10.5 & -- \\
 & Qwen3-8B & 7.2 & 1.4 & $-$28.2 & -- & $+$60.9 & -- \\
 & Qwen3-8B + GRPO & 10.4 & 0.5 & $-$27.3 & -- & $+$57.2 & -- \\
 & Veri-R1 & 22.3 & 5.4 & $-$13.2 & -- & $+$16.9 & -- \\
\addlinespace[2pt]
CONFACT N & Qwen3-4B & 5.9 & 3.1 & $-$3.0 & $+$2.0 & $+$2.6 & $-$2.9 \\
 & Qwen3-4B + GRPO & 4.4 & 1.7 & $-$2.8 & $+$0.8 & $+$2.1 & $-$0.6 \\
 & Qwen3-8B & 4.9 & 3.1 & $-$4.6 & $+$2.4 & $+$3.1 & $-$2.9 \\
 & Qwen3-8B + GRPO & 6.7 & 2.5 & $-$4.7 & $+$2.4 & $+$3.9 & $-$2.9 \\
 & Veri-R1 & 34.6 & 2.2 & $-$23.0 & $+$16.8 & $+$33.1 & $-$17.6 \\
\addlinespace[2pt]
CONFACT P & Qwen3-4B & 8.5 & 1.8 & $-$13.0 & $+$2.3 & $+$13.7 & $-$3.4 \\
 & Qwen3-4B + GRPO & 8.2 & 1.8 & $-$13.7 & $+$3.2 & $+$18.0 & $-$1.5 \\
 & Qwen3-8B & 10.2 & 2.5 & $-$23.6 & $+$6.7 & $+$62.0 & $-$23.1 \\
 & Qwen3-8B + GRPO & 14.1 & 1.7 & $-$24.0 & $+$8.4 & $+$60.7 & $-$29.1 \\
 & Veri-R1 & 50.0 & 7.3 & $-$26.8 & $-$11.0 & $+$41.9 & $+$20.0 \\
\bottomrule
\end{tabular}
\end{adjustbox}
\end{table}

\section{Accuracy in all settings and pilot results}
\label{app:breakdown}

\begin{table}[H]
\centering
\small
\caption{Label accuracy (\%) on the original evidence at 4B, mean over seeds, in all six settings.}
\label{tab:accuracy}
\begin{adjustbox}{max width=\textwidth}\begin{tabular}{@{}lcccccc@{}}
\toprule
Method & AVeriTeC N & AVeriTeC P & SciFact N & SciFact P & CONFACT N & CONFACT P \\
\midrule
GRPO baseline & 61.7 & 57.1 & 71.9 & 69.5 & 71.6 & 64.5 \\
\midrule
Trust-aware reward & 62.5 & 56.8 & 71.5 & 69.1 & 71.5 & 63.8 \\
Trust-aware + abstention & 63.4 & 58.4 & 72.1 & 68.7 & 71.8 & 64.7 \\
\midrule
Consistency, $\lambda{=}0.1$ & 61.8 & 56.4 & 72.3 & 70.2 & 71.9 & 65.6 \\
Consistency, $\lambda{=}0.2$ & 61.4 & 56.9 & 71.8 & 69.8 & 72.1 & 65.0 \\
Consistency, $\lambda{=}0.4$ & 61.8 & 57.6 & 72.5 & 69.8 & 72.0 & 67.0 \\
\midrule
\rowcolor{rowgray} \textbf{TSA (ours)} & 62.3 & 57.4 & 72.3 & 69.7 & 72.1 & 65.7 \\
\rowcolor{rowgray} TSA + consistency, $\lambda{=}0.2$ & 61.8 & 56.3 & 71.5 & 70.0 & 71.4 & 64.8 \\
\rowcolor{rowgray} TSA + consistency, $\lambda{=}0.4$ & 61.2 & 56.8 & 71.8 & 69.3 & 71.7 & 65.9 \\
\bottomrule
\end{tabular}
\end{adjustbox}
\end{table}

\begin{table}[H]
\centering
\small
\caption{Per-class accuracy (\%, mean $\pm$ s.d.\ over seeds) at 4B in the main setting, on the original and the swapped packs. No class collapses under TSA.}
\label{tab:per-class}
\begin{adjustbox}{max width=\textwidth}\begin{tabular}{@{}lcccccc@{}}
\toprule
 & \multicolumn{3}{c}{Original pack} & \multicolumn{3}{c}{Swapped pack} \\
\cmidrule(lr){2-4}\cmidrule(lr){5-7}
Method & \SUP{} & \REF{} & \NEI{} & \SUP{} & \REF{} & \NEI{} \\
\midrule
GRPO baseline & $49.2\pm3.5$ & $69.7\pm5.1$ & $66.2\pm4.9$ & $45.2\pm3.7$ & $69.6\pm4.8$ & $68.5\pm5.4$ \\
TSA (ours) & $50.0\pm1.5$ & $70.5\pm1.5$ & $65.1\pm1.6$ & $46.5\pm1.6$ & $70.3\pm1.9$ & $66.9\pm2.0$ \\
TSA + consistency, $\lambda{=}0.4$ & $49.4\pm3.0$ & $68.1\pm4.2$ & $67.3\pm2.1$ & $45.8\pm3.0$ & $67.8\pm3.1$ & $69.4\pm2.8$ \\
\bottomrule
\end{tabular}
\end{adjustbox}
\end{table}

\paragraph{Flip counts.}
TSA has slightly more confident pairs than the baseline (842 vs.\ 807 on average), so its lower flip rate also corresponds to fewer flips in absolute terms (65.7 vs.\ 71.8 per run). We report rates because the number of confident pairs varies across seeds (739--916 for the baseline).

\paragraph{Trust-aware rewards in single-seed pilots.}
In our own single-seed pilot runs, the two trust-aware rewards appeared to reduce label-driven verdict changes and, for the abstention variant, confident verdicts on weakly supported claims (by more than 10\pp{} in one seed). With five seeds, neither effect holds: Table~\ref{tab:breakdown} shows no change in the flip rate, and the abstention variant changes the rate of confident verdicts on weakly supported claims by $+2.0$\pp{} ($[-6.8,+8.6]$) on AVeriTeC and $+2.9$\pp{} ($[-3.4,+7.9]$) on SciFact relative to the trust-aware reward. This is why all comparisons in this paper use multiple seeds.

\section{Persistent failures}
\label{app:claims}

We group the 984 AVeriTeC claims (main setting) by how many of the five baseline runs flip their verdict. Of the 29 claims that flip in at least four baseline runs, 19 are high-trust packs, 5 are low-trust packs, and 5 are mixed packs; 24 flip mainly through abstention and 5 through reversal. TSA flips 23 of them in all three runs and 6 in one or two runs; TSA + consistency flips 15 in all three runs, 13 in one or two, and 1 in none. Because the total number of flips falls under TSA (65.7 vs.\ 71.8 per run) while these persistent claims keep flipping, TSA's gain comes from claims whose verdict flips in only some runs. A per-group comparison of flip counts would be biased by regression to the mean, since claims are grouped using the baseline runs, so we report only this aggregate conclusion.

\section{Sensitivity to the confidence threshold}
\label{app:threshold}

\begin{table}[H]
\centering
\small
\caption{Change in flip rate (pp) relative to the 4B GRPO baseline in the main setting when confident pairs are defined with different thresholds on $\max(p,p')$ (0: all pairs). Pairs: mean number of pairs kept for the baseline. $\star$: 95\% interval excludes zero.}
\label{tab:threshold}
\begin{adjustbox}{max width=\textwidth}\begin{tabular}{@{}lcccccc@{}}
\toprule
Threshold & Pairs & Baseline (\%) & TSA & TSA + cons.\ ($\lambda{=}0.4$) & Trust-aware reward & Consistency ($\lambda{=}0.4$) \\
\midrule
0.0 & 984 & 7.32 & $-$0.65 & $-$0.91\sig & $+$0.35 & $-$0.07 \\
0.5 & 841 & 8.56 & $-$1.07\sig & $-$1.44\sig & $+$0.59 & $+$0.98 \\
0.6 & 828 & 8.70 & $-$1.03\sig & $-$1.54\sig & $+$0.80 & $+$0.92 \\
0.7 & 807 & 8.92 & $-$1.12\sig & $-$1.60\sig & $+$0.76 & $+$0.79 \\
0.8 & 683 & 10.06 & $-$1.21 & $-$1.51\sig & $+$0.48 & $-$0.55 \\
0.9 & 622 & 8.32 & $-$0.38 & $-$0.60 & $+$1.02 & $-$0.12 \\
\bottomrule
\end{tabular}
\end{adjustbox}
\end{table}

TSA's estimate is negative at every threshold; at 0.8 and 0.9 fewer pairs remain and the intervals widen. Unlabeled packs, which the swap does not change, contribute no flips; excluding them raises the baseline flip rate from 8.92\% to 9.20\% and leaves TSA's reduction unchanged ($-1.14$\pp\sig).

\section{Unseen perturbations in detail}
\label{app:unseen}

\begin{table}[H]
\centering
\small
\caption{Flip rates and flip types at 4B in the main setting under the trained-on swap and the two unseen perturbations. Reversal and abstention are percentages of confident pairs. $\star$: 95\% interval excludes zero.}
\label{tab:unseen}
\begin{adjustbox}{max width=\textwidth}\begin{tabular}{@{}llccccc@{}}
\toprule
Perturbation & Method & Pairs & Flip (\%) & Reversal (\%) & Abstention (\%) & $\Delta$ flip vs.\ baseline \\
\midrule
Label swap (trained on) & GRPO baseline & 807 & $8.92\pm1.11$ & 1.93 & 6.99 & -- \\
 & TSA (ours) & 842 & $7.80\pm0.15$ & 1.91 & 5.89 & $-$1.12\sig [$-$2.04, $-$0.32] \\
 & TSA + consistency, $\lambda{=}0.4$ & 865 & $7.32\pm1.13$ & 1.71 & 5.61 & $-$1.60\sig [$-$2.90, $-$0.18] \\
\midrule
Downshift (unseen) & GRPO baseline & 795 & $4.55\pm0.99$ & 1.37 & 3.18 & -- \\
 & TSA (ours) & 829 & $3.98\pm0.57$ & 1.25 & 2.73 & $-$0.57 [$-$1.52, $+$0.36] \\
 & TSA + consistency, $\lambda{=}0.4$ & 854 & $4.97\pm0.38$ & 1.24 & 3.73 & $+$0.42 [$-$0.45, $+$1.27] \\
\midrule
Label removal (unseen) & GRPO baseline & 810 & $7.86\pm0.59$ & 1.99 & 5.87 & -- \\
 & TSA (ours) & 845 & $6.51\pm0.56$ & 1.68 & 4.83 & $-$1.35\sig [$-$2.01, $-$0.72] \\
 & TSA + consistency, $\lambda{=}0.4$ & 874 & $6.43\pm0.79$ & 1.53 & 4.90 & $-$1.43\sig [$-$2.18, $-$0.50] \\
\bottomrule
\end{tabular}
\end{adjustbox}
\end{table}

Under both unseen perturbations, TSA has fewer reversals and fewer abstention flips than the baseline (point estimates). The downshift changes fewer labels than the swap, so all flip rates are lower. The extra flips of TSA + consistency under the downshift are abstention flips (3.73\% vs.\ 3.18\% for the baseline).

\section{Qwen3-8B details}
\label{app:8b}

\begin{table}[H]
\centering
\small
\setlength{\tabcolsep}{4pt}
\caption{Qwen3-8B in the main setting. Top: the unchanged 4B recipe (learning rate $4\times10^{-5}$) makes training unstable, so we re-selected the learning rate on the baseline arm only. Bottom: the selected recipe ($10^{-5}$), five matched seeds per arm.}
\label{tab:8b}
\begin{adjustbox}{max width=\textwidth}\begin{tabular}{@{}lccccc@{}}
\toprule
Arm & Accuracy (\%) & \NEI{} rate (\%) & Flip rate (\%) & $\Delta$ vs.\ baseline [95\% CI] & Reversal / abstention (\%) \\
\midrule
\multicolumn{6}{@{}l}{\emph{Unchanged 4B recipe (learning rate $4\times10^{-5}$)}} \\
GRPO baseline & $59.1\pm5.1$ & $32.4\pm13.8$ & $16.6\pm5.5$ & -- & -- \\
TSA & $51.2\pm6.0$ & $45.9\pm11.0$ & $27.0\pm5.4$ & -- & -- \\
TSA + consistency & $52.5\pm5.2$ & $45.7\pm8.0$ & $27.1\pm3.4$ & -- & -- \\
\midrule
\multicolumn{6}{@{}l}{\emph{Selected recipe (learning rate $10^{-5}$), five matched seeds}} \\
Untrained Qwen3-8B & $67.0$ & $16.0$ & $10.5$ & -- & $4.2$ / $6.3$ \\
GRPO baseline & $65.1\pm0.6$ & $26.0\pm1.0$ & $15.0\pm1.2$ & -- & $2.6$ / $12.5$ \\
TSA & $64.2\pm0.8$ & $24.7\pm1.7$ & $14.7\pm1.6$ & $-$0.36 [$-$2.21, $+$1.33] & $3.0$ / $11.7$ \\
TSA + consistency & $64.4\pm0.3$ & $24.1\pm1.3$ & $14.3\pm1.3$ & $-$0.74 [$-$2.49, $+$0.58] & $3.2$ / $11.1$ \\
\bottomrule
\end{tabular}
\end{adjustbox}
\end{table}

\begin{table}[H]
\centering
\small
\setlength{\tabcolsep}{4pt}
\caption{Qwen3-8B in all six settings (selected recipe). Flip rates in \%; $\Delta$ columns are matched-seed differences in pp. GRPO training raises the flip rate above the untrained model in every setting (all five seeds). Neither intervention changes the flip rate significantly in any setting. The only significant method-versus-method contrast is TSA + consistency minus TSA on CONFACT with the principled prompt ($-1.19$, $[-1.90,-0.43]$), a different setting from the one significant contrast at 4B; we treat it as exploratory.}
\label{tab:8b-cells}
\resizebox{\textwidth}{!}{\begin{tabular}{@{}lcccccc@{}}
\toprule
Setting & Untrained & GRPO baseline & TSA & TSA + cons. & $\Delta$ TSA & $\Delta$ TSA + cons. \\
\midrule
AVeriTeC N & 10.5 & $15.0\pm1.2$ & $14.7\pm1.6$ & $14.3\pm1.3$ & $-$0.36 & $-$0.74 \\
AVeriTeC P & 15.8 & $22.6\pm0.4$ & $22.6\pm0.8$ & $22.3\pm0.8$ & $+$0.05 & $-$0.30 \\
SciFact N & 6.2 & $12.7\pm1.6$ & $13.1\pm1.1$ & $13.4\pm1.0$ & $+$0.33 & $+$0.64 \\
SciFact P & 7.2 & $10.4\pm0.9$ & $10.7\pm2.2$ & $10.2\pm2.4$ & $+$0.30 & $-$0.18 \\
CONFACT N & 4.9 & $6.7\pm1.0$ & $6.8\pm0.5$ & $7.1\pm0.7$ & $+$0.07 & $+$0.39 \\
CONFACT P & 10.2 & $14.1\pm1.3$ & $14.7\pm0.9$ & $13.5\pm0.6$ & $+$0.57 & $-$0.62 \\
\bottomrule
\end{tabular}}
\end{table}

\paragraph{Confidence response at 8B.}
At 8B, TSA does not weaken the confidence response relative to the baseline ($+0.54$, $[-0.11,+1.16]$), while TSA + consistency has a weaker response than TSA ($-0.50$\sig, $[-0.83,-0.24]$, five of five seeds), in line with our recommendation of plain TSA.

\end{document}